\documentclass{article} 
\usepackage{iclr2027_conference,times}

\usepackage{amsmath,amsfonts,bm}

\def\eqref#1{equation~\ref{#1}}

\def\1{\bm{1}}

\DeclareMathAlphabet{\mathsfit}{\encodingdefault}{\sfdefault}{m}{sl}
\SetMathAlphabet{\mathsfit}{bold}{\encodingdefault}{\sfdefault}{bx}{n}

\usepackage{hyperref}
\usepackage{url}
\usepackage{subcaption}
\usepackage{multirow}
\usepackage{rotating}
\usepackage{booktabs}

\title{Local Temporal Contrasts on Human Motion Data Induce a Latent Geometry}

\author{Lourenço Abrunhosa Rodrigues \& Felix Putze \\
Cognitive Systems Lab\\
FB3, Department of Mathematics and Computer Science, University of Bremen\\
Bremen, 28359 Bremen, Germany \\
\texttt{\{lourenco, fputze\}@uni-bremen.de} \\
}

\iclrfinalcopy 
\begin{document}

\maketitle

\begin{abstract}

Time series such as recordings of human motion data often encode structure across multiple temporal scales, and when a latent state determining this sequence evolves, this structure needs to survive in the representation space for downstream use. Yet the currently prevalent attention-based encoders paired with sequence-level pooling have made single-vector, pointwise comparison the default mode of inference, discarding the trajectory a sequence traces through the representation space. We propose recovering this geometric structure directly, without relying on forecasting-based objectives, by training an encoder model (SemTAC) with a temporal-neighborhood contrastive loss: representations that satisfy local consistency everywhere must give rise to a coherent global structure, letting inference happen at temporal scales beyond those seen during training. We evaluate SemTAC against four baselines on a five-step motion sequence discrimination task: the same architecture trained with a classification or a standard contrastive objective, and two established self-supervised time-series methods, TS-TCC and TS2Vec. At the full sequence length all five are broadly competitive, but on incomplete, truncated sequences SemTAC holds a large advantage: a median F1 of 0.77 one step before the sequence completes, against 0.27–0.70 for the four baselines. A hyperparameter-matched ablation attributes this effect specifically to the temporalcontrastive objective. SemTAC's representation trajectories are also the only ones among these five to exhibit interpretable geometric structures, exceeding a no-structure chance baseline on a displacement-smoothness metric which we introduce. This is evidence that its objective, not classification accuracy alone, produces genuinely temporally-ordered trajectories.

\end{abstract}

\section{Introduction}

Capturing the temporal structure of data has long been a focus of time-series research, and classical models such as ARIMA as well as modern architectures such as recurrent networks and Transformers have made substantial progress in this direction. In domains such as biosignals, this has enabled applications ranging from medical diagnosis to activity recognition. Two challenges, however, still limit how much temporal structure these models retain for downstream use: identifying the right timescale for analysis, and preserving how the latent state evolves once that timescale is chosen. This latent state is what conditions the observed signals, similar to how the intended action of a subject conditions their physical movements, or the progression of a disease changes the symptoms that are presented.

Biosignals often encode information across multiple temporal scales, from fast, beat-to-beat variability to slower trends such as disease progression. Existing approaches typically address this by fixing a single temporal resolution or aggregating information over time, which limits their ability to capture dynamics that occur at other scales.

A separate issue concerns what happens to the representation once a scale is chosen. Recurrent architectures were built around an evolving state, representing a sequence as a trajectory rather than a single point. Attention-based architectures shifted this convention: they aggregate a sequence directly and are typically paired with a pooling step that collapses it into a single vector, so inference reduces to comparing or clustering points and discards the structure of how those points were reached. Such representations can say that two states are similar, but not where a given state lies within a process, or what is likely to follow it.

This matters especially when states evolve in stages rather than continuously. Many approaches to temporal representation rely on forecasting (e.g. \citet{eldele_time-series_2021}), assuming a signal's own dynamics are enough to extrapolate what comes next: an assumption that breaks down when, for instance, a health condition advances through the appearance of a qualitatively new symptom rather than the gradual evolution of existing ones.

We propose that this trajectory structure can be induced directly, without forecasting, via contrastive learning approaches such as Temporal Neighborhood Coding \citep{tonekaboni_unsupervised_2021}, which contrasts temporally proximal states against distant ones. A model trained to correctly capture these local interactions must give rise to a global structure that supports them everywhere along the trajectory, not only at points seen during training.

This global structure lets us compose individually-encoded states into longer trajectories and compare them geometrically at inference, without retraining at a coarser scale and directly addressing the first challenge. It is also useful for early prediction and monitoring, e.g. of an individual's health state, since it lets us identify which states precede others in the representation space, not only distinguish between labels.

\begin{figure}
    \centering
    \includegraphics[height=4.5cm]{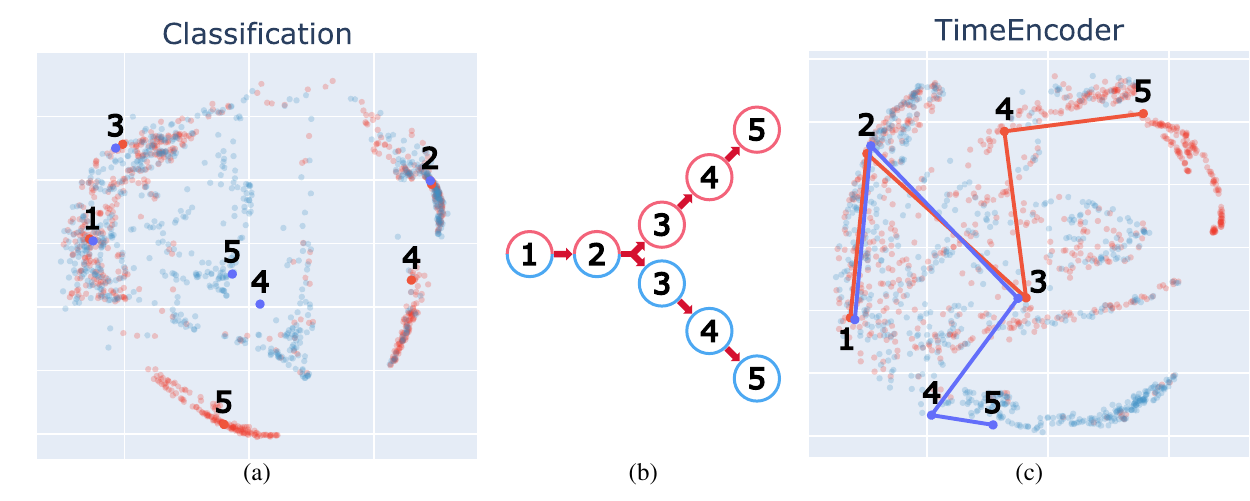}
    \caption{Training our model with temporal contrasts induces a structure in the latent space that mirrors the temporal structure of the data. (a) shows the representation space when trained with purely discrimination-centric objective, (b) shows the original data structure (see Fig. \ref{fig:sequence_structure}), and (c) is the latent space created by our approach.}
    \label{fig:goal-illustration}
\end{figure}

Figure \ref{fig:goal-illustration} illustrates the perspective shift this work introduces: Instead of learning representations that hold isolated clusters of defined categories, induce a geometric structure over which the temporal structure of the dataset is represented. That structure can then be used to identify what "type" of sequence a new set of samples belongs to. Finally, if finer scales correspond to their own classes, it is possible to recover those as granular components of the overall structure.

Our contributions are as follows:

\begin{itemize}
    \item We introduce a geometry-based perspective on representation learning for time series data, where they are modeled as structures in latent space rather than static embeddings, and portions of those structures remain informative for instantaneous state classification.
    \item We show that contrastive learning with temporally-informed transformations induces structured latent dynamics, linking local temporal constraints to global geometric organization in representation space.
    \item We propose geometry-level inference via sequence similarity measures applied in latent space, enabling comparison of entire representation geometries rather than individual points.
    \item We empirically demonstrate that this approach improves robustness to partial sequences, consistently outperforming contrastive and classification baselines in sub-sequence-based inference tasks.
    \item We measure structural properties of the learned representation spaces and show that our contrastive approach based on local temporal interactions results in significantly better structured spaces.  
\end{itemize}

\section{Related Work}


Contrastive learning has emerged as a dominant paradigm for representation learning. Introduced by \citet{oord_representation_2019} under the name Contrastive Predictive Coding, it formulates representation learning as mutual information maximization between positive pairs in contrast to negative samples. Subsequent work, particularly in computer vision, has demonstrated its effectiveness, including \citet{chen_simple_2020}, \citet{he_momentum_2020}, and \citet{pmlr-v139-radford21a}.

A theoretical perspective is provided by \citet{wang_understanding_2020}, who characterize contrastive objectives in terms of alignment and uniformity. In practice, these methods rely on data augmentations to construct positive pairs, thereby shaping the invariances of the learned representation space (\citet{von_kugelgen_self-supervised_2021}). However, these approaches operate primarily on static data and do not explicitly account for temporal dependencies in sequential settings.


Learning representations of sequential data has a long history, with early approaches relying on recurrent architectures such as Long Short-Term Memory (\citet{hochreiter_long_1997}) and sequence-to-sequence models (\citet{sutskever_sequence_2014}). More recently, Transformer-based architectures (\citet{vaswani_attention_2017}) have become the standard for modeling long-range dependencies in sequences.

A common characteristic of these approaches is the use of temporal aggregation mechanisms such as averaging, max-pooling, or attention-based pooling (e.g. \citet{yang_hierarchical_2016}) to produce fixed-dimensional representations of variable-length sequences. While effective for downstream tasks, such aggregation compresses temporal dynamics into a single vector, potentially discarding information about the evolution of representations over time.


Recent work has extended contrastive learning to time series by leveraging domain-specific properties of sequential data. These include local temporal consistency, \textit{i.e.}, the assumption that temporally proximal segments correspond to similar latent states, and transformation invariance, \textit{i.e.}, robustness to meaning-preserving transformations. Methods such as \citet{yue_TS2vec_2022, foumani_series2vec_2024, eldele_time-series_2021, tonekaboni_unsupervised_2021} demonstrate the effectiveness of these principles across tasks including classification, forecasting, and anomaly detection. \citet{tonekaboni_unsupervised_2021} in particular applies a temporal contrast very similar to ours, with the main difference that we define pairs across stages (different motions) that would not be considered by their approach.

Despite these advances, existing approaches do inference at the level of individual representations, either by comparing point embeddings or clustering them. Although multi-scale representations have been explored (\citet{yue_TS2vec_2022}), the structure induced by temporal consistency is not explicitly leveraged at the level of trajectories in representation space.


A large body of work leverages the geometry of representation spaces for downstream tasks. This is ultimately the goal of Deep Metric Learning (DML) (\citet{kaya_deep_2019}), which encodes sample relationships as distances in a representation space to obtain relevant features. Contrastive Learning is one of the approaches used in this field.

Beyond static embeddings, cluster-quality metrics have been adapted to trajectory-structured data in other domains, \textit{e.g.}, a silhouette-based evaluation of longitudinal clinical clustering (\citet{luong_longsil_2019}) and deep trajectory phenotyping of ICU time series (\citet{wu_deep_2024}). To our knowledge, however, no prior work directly characterizes the branch-like or temporal-order structure of learned representation trajectories themselves, a gap the metrics introduced in Section~\ref{sec:branch_quality} address.

\section{Methods}

\subsection{Dataset Preparation}
\label{sec:dataset-preparation}

The data used in this work are prepared to provide a strong ground truth on the temporal structure of the data. To do so, we employ a generative model, the Motion Diffusion Model (MDM) (\citet{tevet_human_2023}), to synthesize large amounts of motion data with well known activity labels. We choose to use motion data for this work because it is a complex multi-channel time-series data modality that requires non-trivial representation learning to distinguish what actions are being performed; and it can plausibly be used to model our scenario where internal states (in this case the type of motion being performed) are encoded in a time-series which behavior does not directly inform on future states. We define 10 types of motion as prompts with the structure \verb+ "A person <predicate>." + and create 20 rephrasings of each using ChatGPT (o3-mini). For each prompt we synthesize 100 motion instances and filter them using the method proposed by \citet{biosignals26}. For full details on dataset preparation, see Appendix \ref{app:generation_details}. 

From the filtered motions we create sequences of motions that hold a temporal structure that our method should learn to represent. Of the 10 types of motion, two were excluded from sequence creation as a consequence of the filtering results.  This data is designed as sequences of arbitrary motions to avoid that the models simply learn some physical dynamics of signal progression, and instead are forced to model the global temporal structure purely based on the conditioning that the contrastive learning framework induces.

\begin{figure}
    \centering
    \includegraphics[width=.8\linewidth]{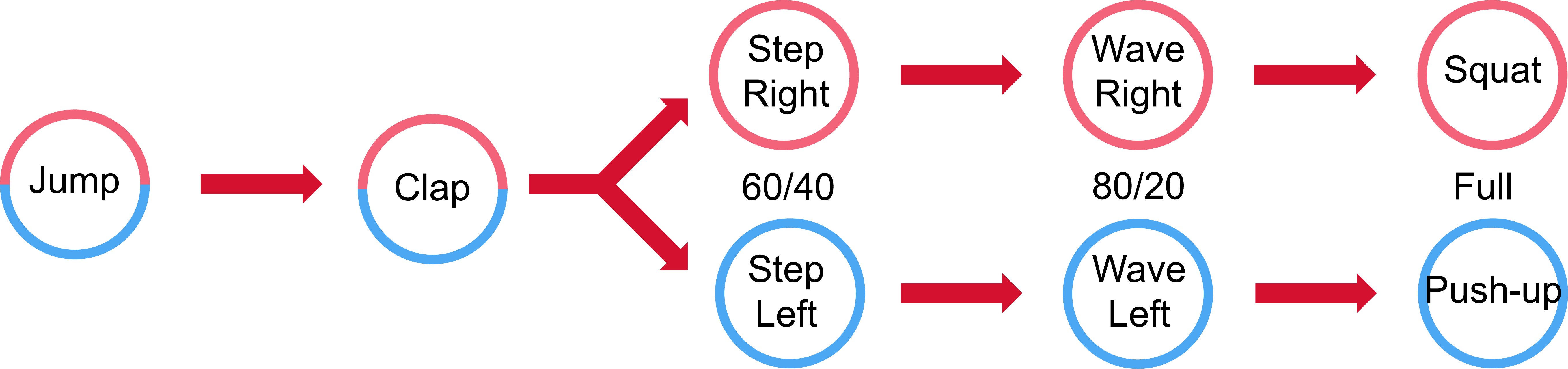}
    \caption{Generated motion sequences with divergent structure. Both sequence types begin identically (jump, step), then diverge toward a preferential lateralization before performing completely unrelated motions. Step Right/Left labels are 60/40\% corrupted (only 60\% match the stated direction); Wave Right/Left labels are 80/20\% corrupted.}
    \label{fig:sequence_structure}
\end{figure}

The structure we evaluated in this work is a simple divergent structure, as can be seen in Figure \ref{fig:sequence_structure}. The first two steps of both sequences are shared, and after that the sequences become evermore lateralized, until the final sequence step performs completely different motions (push-ups versus squats). To make the lateralization gradual, and to prevent sequence type from being trivially identifiable from the first non-shared step, we introduce a mixing factor into the lateralized motions (step right/left, wave right/left). Rather than each sequence containing a purely one-sided lateralization, we assign each sequence a dominant lateralization and relabel a fraction of its instances to the opposite side: 40\% for the step motions and 20\% for the wave motions. This gives each sequence only a slight lateral majority rather than a pure one.

This process results in 1420 unique sequences: 710 with a right-lateralized majority and 710 with a left-lateralized majority, which we refer to as \textit{the right sequence} and \textit{the left sequence}, respectively.

This dataset is divided into train-validation-test splits by assigning whole sequences to each split, such that 60\% are in training, 20\% in validation and 20\% in test. The same splits are used for all models.

\subsection{Temporal Contrasts}

To encourage the model to learn the temporal structure of these sequences, we use a contrastive learning framework \citep{oord_representation_2019}. To do so, we design the transformations that produce positive pairs to reflect the intuitive temporal relationships that the data should hold.
What we refer to as a sample is a 6 s time segment of a given motion instance.
We define the following two transformations:
\begin{itemize}
    \item \textbf{Partial overlap:} if two samples are taken from the same motion instance and have partial overlap of their corresponding time-series, then they are a positive pair (self-similarity).
    \item \textbf{Sequence adjacency:} if two samples are taken from motion instances that are adjacent to each other in a given sequence, then they are a positive pair (local temporal interaction).
\end{itemize}

 To efficiently implement this, we created a python package \textit{contrastlearn}\footnote{ \url{https://anonymous.4open.science/r/ContrastLearn-C97C/} } that defines samples as data structures that point to the segment to be read-out from a file and includes information about the sequence and motion instance to which the sample belongs. Then finding partial overlap or sequence adjacency positive pairs for any given sample corresponds to simply adjusting the read-out starting timestamp or the sample index in the sequence.

At each training step, a batch of anchors and their corresponding positive samples are retrieved. A positive sample is created for each anchor, randomly selecting one of the two transformations to be applied. In our experiments, that selection followed a uniform distribution. Representations are obtained by passing anchors and positive samples through the model, and these representations are then used to compute the InfoNCE loss. Negatives are sampled from the batch of anchors as the set excluding the anchor itself.

\subsection{Model Architecture}

In this work, we use the Semantic Encoder model from \citet{Richardson2025}. This corresponds to a transformer encoder model suitable for learning representations of motion data in the same structure as that of the motion instances generated for this paper. We combine the Semantic Encoder architecture with a TNC-like objective in which motion adjacency replaces the temporal neighborhood, and call the resulting model Semantic Temporal Adjacency Coding (SemTAC).
The architecture of this model is unchanged, with the exception of the output dimension, which is adjusted to the desired representation dimension in our tests, and the addition of a linear layer with that same dimension to have a final linear projection step as is common in contrastive learning.
In all of our tests, the model is randomly  initialized.
Model and training hyper-parameters are defined through a grid search, selecting a range of parameters that result in well behaved training and validation loss curves, \textit{i.e.} visible L-shaped curves within 10 epochs of running.

The parameters evaluated in this grid search were batch size $=\{8, 16, 32, 64\}$, learning rate $=\{10^{-2}, 10^{-3}, 10^{-4}, 10^{-5}\}$, dimension of the representation space $=\{2, 10, 32, 64\}$ and the temperature factor of the InfoNCE loss $=\{0.05, 0.1, 0.3, 0.5\}$. All models were trained on an RTX 6000 GPU (CUDA 12.04) and ADAM optimizer until they reached 100 epochs or when early stopping triggered after a patience interval of 20.

\subsection{Evaluation}
\label{sec:branch_quality}

We want to evaluate whether the model is capable of learning the temporal structure of the dataset and mirroring it in its representation space. Our hypothesis is that to learn this structure, the latent space has to structure itself around the temporal trajectories that the data were designed to follow. To measure if this truly occurs, we measure two things: (1) if evaluating the representation geometry of test sequences (and sub-sequences) allows us to correctly classify to which sequence type (left or right) they belong; and (2) representation space properties that reflect the space's geometric structure and ability to separate the two sequence types.

To tackle the first, we introduce a geometric inference method, which compares the geometric structures of known prototype patterns against those of new queries. We do so by borrowing sequence/trajectory distance methods, Dynamic Time Warping (DTW) \citep{sakoe_dynamic_1978} and Fréchet distance \citep{alt_computing_1995}. After model training, we used the validation set to establish an average prototype for the left and right sequences (computed as the sequence of centroids of each individual step). DTW and Fréchet distances between these prototypes and each of the query sequences are computed using the implementations provided by \citet{Jekel2019}, and converted to probabilities using an RBF kernel method, where the kernel width is estimated as the inter-prototype distance. We evaluate both full-sequence classification performance, as well as performance at step 4. This tests the capability of the model to leverage the cumulative information of knowing what step 3 and step 4 are, but, given their label noise, does not provide the model with a fully confident anchor on which to base the final prediction.

The second set of measurements consists of two groups of metrics:
\begin{itemize}
 \item \textbf{Temporal encoding} - we measure how well time is represented as a direction in the representation space, for this we measure what we call smoothness  and temporal order recoverability. Smoothness is the average cosine similarity between the displacement vectors between consecutive pairs of points in the representation space, if time is a driving factor in the representation space, we expect that a sequence as its points aligned along a preferential direction (and so smoothness values closer to 1). Given that this method applies the equivalent to a autocorrelation with lag 1, the baseline value for this metric is actually $-0.5$ and not zero as we would naively expect. (See Appendix \ref{smoothness-baseline-explanation} for detailed explanation). Temporal order recoverability corresponds to fitting the sequence representations for a given sequence type to a monotonically increasing line. To have a good fit, the representation space needs to have all sequences of a given type agree on a progression direction, and keep their individual steps in an order compatible with the original motion order.
 \item \textbf{Sequence branching} - we measure how well the representation space separates the two types of sequence, independently of prototype construction. For this we measure the silhouette score \citep{rousseeuw_silhouettes_1987} and a KNN-based consistency metric. The first measures the ratio between average intra- and inter-class (sequence type) distances between the test set sequences, while the consistency metric measures if, on average, the closest neighboring sequence is a sequence of the same type.
\end{itemize}

To test the statistical significance of the multiple comparisons between our SemTAC approach and the other models, we first perform an omnibus test using the Kruskal-Wallis test, and, after a significant result, we perform Mann-Whitney U tests between our approach and each of the others.  Given the multiple tests that are performed for each observation, we apply the Holm-Bonferroni correction, using 0.05 as the base significance level.

The previous tests focus on the global temporal scale of the data. To evaluate the retention of information at a local scale we do single motion classification. We use the validation data to train a logistic regression classifier on the latent features to classify the type of motion each sample corresponds to, reporting the macro F1-score over the 8 motion classes.

Finally, we visualize the representation spaces created by the trained models. To do so, we produced PHATE plots \citep{moon_visualizing_2019}, where all representations of the test points, as well as the coordinates of the patterns in the representation space, are used to fit the transformation. The selection of which model instances for each approach to render in these plots was based on a balanced score of model performance and representation space organization. For each available model, the minimum between its F1 score at full sequence and the average of the 4 space metrics (Temporal Encoding and Sequence Branching ones) is chosen, and the run presenting the highest balanced score is chosen to represent the approach. This way, we make sure that the selected models perform their best on both dimensions.

\textbf{Baseline models.} As comparison baselines, we use TS-TCC and TS2Vec as these are still the most prominent models when it comes to time series representation learning using contrastive learning. Both are contrastive learning-based models especially designed to be general time series encoders. Both models produce a representation vector for each timestamp in the input time series, so we apply a specific time aggregation method to each of them so that their output matches the 5-point sequence shape of our experiment.

TS-TCC creates a strong and weak augmentation and uses one to predict future observations of the other, and \textit{vice-versa} (temporal contrast), and contrasts the context representations (summary of sequence) of corresponding sequences against others (context contrast). The weak and strong transformations are perturbations to the original signal (jitter and scaling, and segment-order permutation). During training, TS-TCC learns a context head that summarizes the contests of a set of timestamp representations into a single embedding point. We recover this head to aggregate the outputs of this model into our 5-point sequences, so we create a motion specific context for each motion instance.

TS2Vec pairs the corresponding timestamps in two views of the same sequence against different timestamps of the same sequence (temporal contrast), and against samples at the same timestamp from other sequences (instance contrast). Positive pairs are built by matching the corresponding timestamps and masking sections of one of the sequences. As already done in \citet{yue_TS2vec_2022}, we use max pooling to aggregate the timestamp embeddings into the coarser time-scale in which our data exist.

Finally, to give these models the best chance in terms of global temporal structure learning, we feed them full 5-motion sequences, which are produced by concatenating the motion instances that compose each sequence.

\textbf{Ablations.} To provide a direct evaluation on the effects of our approach, we define two ablations, keeping the same model architecture, but changing the training strategy:
\begin{itemize}
    \item \textbf{Self-similarity only} - we deactivate the sequence adjacency transformation, which should remove the models ability to learn the temporal structure of the data. In the rest of the paper we refer to the models trained with this approach as "Self-similarity".
    \item \textbf{Supervised training} -  we train the model to predict the types of motion that correspond to the samples being processed. In the rest of the paper we refer to the models trained with this approach as "Classification".
\end{itemize}

\section{Results}

\textbf{Performance at full sequence.} We first look at model performance for full sequences. In this scenario, each model has access to the entire 5-motion sequence, and we use this stage as a measure of overall utility of each method, \textit{i.e.} if the models learn any relevant information, then they must be able to distinguish the two sequences at a satisfactory level. These results are shown in Table \ref{tab:full-sequence-performance}, where it is possible to verify that all models produce useful representations, with all models achieving average F1-scores above 0.8 for at least one of the distance methods. 

In the tables presented in this work we use \textbf{bold} to identify the best result in each line, * to mark results significantly different from our full model (SemTAC), and a $^\dagger$ and $^\ddagger$ to identify the baseline and ablation models, respectively.

\begin{table}[htp]
    \centering
    \begin{tabular}{c|ccccc}
        Distance & TS2Vec$^\dagger$ & TS-TCC$^\dagger$ & \textbf{SemTAC} & Self-similarity$^\ddagger$ & Classification$^\ddagger$ \\
        \hline
        Fréchet & $0.81\pm0.04$* & $0.55\pm0.03$* & $0.87\pm0.12$ & $0.83\pm0.15$ & $\mathbf{0.95\pm0.06}$* \\
        DTW & $0.91\pm0.02$* & $0.82\pm0.03$* & $0.86\pm0.06$ & $0.89\pm0.07$ & $\mathbf{0.95\pm0.06}$*
    \end{tabular}
    \caption{F1-scores (mean $\pm$ std) per approach, using Fréchet or DTW as the latent sequence distance.}
    \label{tab:full-sequence-performance}
\end{table}

\textbf{Performance at truncated sequence.} We then evaluate the model performance when the input sequences are truncated to not include the last (and determining) step. Here we hypothesize that the models would have to properly learn the relation between sequence steps to overcome the label noise of steps 3 and 4. By design, our dataset only keeps this relation in the temporal structure, so we expect that a good representation of this structure results in improved performance.
Table \ref{tab:truncated-sequence-performance} shows the mean and standard deviation of the F1-scores obtained in this case. Our model, SemTAC, shows a significant improvement compared to all other models regardless of the distance method used to compare the sequences. The only exception is the Classification ablation using DTW, where our model's higher performance is not statistically significant. The median F1-score for SemTAC is 0.77 (Fréchet) and 0.78 (DTW), close to the theoretical maximum of 0.8 given label noise at step 4. The ~0.1 gap between SemTAC's mean and median F1-score comes from 3 outliers (F1 = 0.00, 0.01, 0.48), all with output dimension 2; the remaining models show Intrinsic Dimension \citep{facco_estimating_2017} consistently between 3 and 4, suggesting these outliers stem from a 2-dimensional space's limited capacity to encode the relevant data properties.

DTW yields higher F1-scores for four of the five approaches. The gains are largest for TS2Vec (0.70 vs. 0.40) and the two ablations, Self-similarity (0.64 vs. 0.37) and Classification (0.63 vs. 0.37), whereas for SemTAC the difference is only 0.04 (0.72 vs. 0.68). TS-TCC is the only approach that scores higher with the Fréchet distance (0.51 vs. 0.47), with both corresponding to the chance level.

\begin{table}[htp]
    \centering
    \begin{tabular}{c|ccccc}
        Distance & TS2Vec$^\dagger$ & TS-TCC$^\dagger$ & \textbf{SemTAC} & Self-similarity$^\ddagger$ & Classification$^\ddagger$ \\
        \hline
        Fréchet & $0.40\pm0.19$* & $0.51\pm0.03$* & $\mathbf{0.68\pm0.21}$ & $0.37\pm0.28$* & $0.37\pm0.32$* \\
        DTW & $0.70\pm0.05$* & $0.47\pm0.12$* & $\mathbf{0.72\pm0.19}$ & $0.64\pm0.15$* & $0.63\pm0.25$
    \end{tabular}
    \caption{F1-scores (mean $\pm$ std) per approach, using Fréchet or DTW as the latent sequence distance. }
    \label{tab:truncated-sequence-performance}
\end{table}

\textbf{Space structure metrics.} In this section, we collect the results related to the different metrics of space structure. As before, we conceptually separate the metrics into two groups. Table \ref{tab:space-metric-temporal-structure} shows the results of the comparison between SemTAC and the other models with respect to the two metrics that measure the temporal structure within the representation (temporal encoding). The SemTAC models are significantly better at both Order Recoverability and smoothness than the others, with the exception of the comparison against the Classification ablation for Order Recoverability where our model still shows a positive advantage but does not cross the significance threshold. For a fair comparison of Order Recoverability, we also measure how much of the data's variance is explained by the linear projection that is correlated with time progression, and verified that (1) SemTAC has the largest values ($0.30\pm0.23$), confirming that the positive results observed are not coincidental or derived of infinitesimal dimensions within the latent space; and (2) that the Classification ablation models have a slightly lower, but still comparable amount of explained variance ($0.24\pm0.18$). Notably, our model is the only that is capable of clearing the -0.5 smoothness value associated with an unstructured sequence, and thus shows a measurable proof of it containing the desired organization around temporal structure.

\begin{table}[htp]
    \centering
    \begin{tabular}{c|ccccc}
        Metric & TS2Vec$^\dagger$ & TS-TCC$^\dagger$ & \textbf{SemTAC} & Self-similarity$^\ddagger$ & Classification$^\ddagger$ \\
        \hline
        Smoothness & $-0.57\pm0.02$* & $-0.53\pm0.01$* & $\mathbf{-0.26\pm0.30}$ & $-0.58\pm0.12$* & $-0.58\pm0.17$* \\
        Order Rec. & $0.67\pm0.04$* & $0.48\pm0.04$* & $\mathbf{0.79\pm0.10}$ & $0.66\pm0.15$* & $0.73\pm0.11$ \\
        (Exp. Var.)& $0.06\pm0.01$ & $0.20\pm0.04$ & $0.30\pm0.23$ & $0.26\pm0.27$ & $0.24\pm0.18$
    \end{tabular}
    \caption{Temporal Encoding results (mean $\pm$ std). Beyond the two metrics described above, we report the explained variability (Exp.\ Var.) of the Order Recoverability metric as validation.}
    \label{tab:space-metric-temporal-structure}
\end{table}

In Table \ref{tab:space-metric-sequence-separation} we compare the SemTAC models with the other approaches in terms of their ability to separate the two types of sequences in the representation space. Similar to our analysis of model performance at full and truncated sequence, while Temporal encoding metrics directly measure the properties we want in the structure of the representation space, Branch Separability metrics verify the utility of said structure. We see that our model is the one that best approaches the separability capacity of the Classification ablation, which is particularly geared towards excelling at these metrics, given that it was specifically trained to isolate the components of each sequence into their own clusters. Despite this positive observation, there is still a significant difference between these two approaches, indicating a trade-off between global structure learning and dedicated component (motion) classification as learning objectives.

\begin{table}[htp]
\centering
\begin{tabular}{lr|ccccc}                             \\
\multicolumn{2}{l|}{Metric}   & TS2Vec$^\dagger$ & TS-TCC$^\dagger$ & \textbf{SemTAC} & Self-similarity$^\ddagger$ & Classification$^\ddagger$ \\ \hline
\multicolumn{2}{l|}{Silhouette}   & & & & & \\
    & Fréchet                     & $0.05\pm0.02$* & $0.02\pm0.01$* & $\underline{0.13\pm0.13}$ & $0.09\pm0.06$* & $\mathbf{0.29\pm0.15}$* \\
    & DTW                         & $0.10\pm0.01$* & $0.09\pm0.02$* & $\underline{0.20\pm0.11}$ & $0.15\pm0.07$* & $\mathbf{0.35\pm0.10}$* \\ \hline
\multicolumn{2}{l|}{kNN-Consist.}   & & & & & \\
    & Fréchet                     & $0.87\pm0.02$* & $0.71\pm0.04$* & $\underline{0.95\pm0.07}$ & $0.91\pm0.08$* & $\mathbf{1.00\pm0.01}$*\\
    & DTW                         & $0.90\pm0.02$* & $0.79\pm0.04$* & $\underline{0.94\pm0.06}$ & $0.91\pm0.06$* & $\mathbf{1.00\pm0.01}$*
\end{tabular}
\caption{Sequence Branching metrics (mean $\pm$ std). \underline{underlined}: second best per distance.}
\label{tab:space-metric-sequence-separation}
\end{table}

In Table \ref{tab:single-motion-performance} we evaluate the models' representation spaces' capacity to support single motion classification. Given the balanced nature of the dataset, and the 8 classes, all models show a high performance compared to the chance level of 0.125, supporting the claim that our approach maintains multi-scale information at different latent structural resolutions. However, all models, except the Self-similarity ablation, show a significantly better performance.

\begin{table}[htp]
    \centering
    \begin{tabular}{ccccc}
        TS2Vec$^\dagger$ & TS-TCC$^\dagger$ & \textbf{SemTAC} & Self-similarity$^\ddagger$ & Classification$^\ddagger$ \\
        \hline
        $\mathbf{0.98\pm0.00}$* & $0.86\pm0.02$* & $0.85\pm0.14$ & $0.80\pm0.23$ & $0.92\pm0.07$* \\
    \end{tabular}
    \caption{F1-scores (mean $\pm$ std) for single motion classification.}
    \label{tab:single-motion-performance}
\end{table}

\begin{figure}
    \centering
    \includegraphics[width=\linewidth]{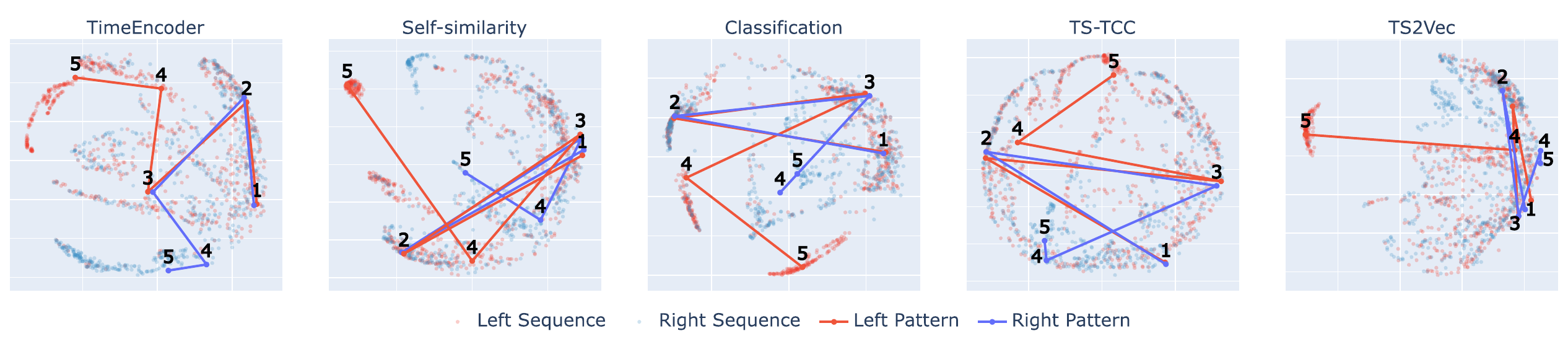}
    \caption{PHATE visualization of the latent space of an example run from each type of model. Points represent actual data points from the test split, while the solid poly-lines represent the prototypes calculated from the validation data (red for left sequence, blue for right sequence). The numbers denote the step positions for each of the sequence prototypes.}
    \label{fig:space-visualization}
\end{figure}

Finally, we show a visualization of a selected run for each type of model in Figure \ref{fig:space-visualization}. There, it is possible to see a clear divergent structure for the SemTAC model that does not exist in any other model. Although not shown here, we have verified that the types of distributions observed in this figure are representative of what is observed for the rest of the runs of each of these types of models.

\section{Discussion}

The results support the main thesis of this work: using local temporal interactions as contrasts induces a structure within the representation space that reflects the global temporal structure of the dataset. From the increased performance for incomplete sequences to the consistent advantages seen for each of the metrics with which we characterized the representation spaces, the SemTAC models show an internal organization around the branching temporal structure that we built for these experiments. Particularly from the comparison with the other contrastive approaches, and especially the Self-similarity ablation, we see that it is the addition of the segment adjacency transformation that drives this behavior. Our contribution is therefore not a new architecture or objective, but the finding that adjacency-based contrasts give rise to representation spaces organized like the underlying temporal structure, and the analysis that isolates the component responsible.

We do see a few points of trade-off: inducing the branching structure costs some isolation of motion-type clusters, which is why the Classification ablation beats SemTAC on metrics that reward clearer motion separation. That cost seems tied less to branching itself than to the self-similarity transformation: a design choice whose effect shows up just as much in the Self-similarity ablation, which lacks branching structure yet shows the same sensitivity in the single-motion classification results, suggesting a better transformation could help both.

Future work will focus on finding methods to control this trade-off, balancing the self-similarity and temporal interaction transformation rates, and introducing more complex and nuanced transformation definitions are avenues that can provide great insight into this.

Regarding sequence comparison methods, DTW is the better choice for our inference approach. We hypothesize that this is because the Fréchet distance depends on the largest pointwise deviation between sequences, which makes it sensitive to outliers such as the mislabeled samples at steps 3 and 4. In the unstructured representation spaces of the other models, such a sample may be embedded far from its motion cluster, whereas SemTAC's organized space would keep it close, which could explain its small gap. Confirming this would require experiments with varying rates of mislabeling. TS-TCC performs near chance level under both distances, so its reversed gap is not meaningful.

In this work, we have explored the structure inducing effect of temporal contrasts using a synthesized dataset. We did so because having full knowledge about the temporal structure of the data is fundamental to evaluate the validity of our observations. Now, with these results as a solid basis, we have to apply the same methodology to real datasets and report what structures can be learned, and so generalize these results into a validated approach for evolving time series modeling.

\subsection*{AI use statement}

(This section is \textbf{required} and does not count toward the page limit.)

In this work, we used generative AI tools for generating the synthetic dataset.
We have not used generative AI tools for developing theoretical models or conceptual frameworks,  propose or refine hypotheses, design or provide feedback on research  methodology or experiments, implement methods, assist with translation, clean and reformat dataset, support qualitative and thematic data analysis, or interpret results. Formulate mathematical claims, provide critical ingredients for proving mathematical claims, and assist in the writing of proofs are not applicable to this work.
Additionally, we used generative AI tools for create and edit software code, and edit the research paper to improve readability. We have reviewed all AI-assisted work. We have used a dataset verification method to assess the quality of the synthesized dataset, all code was reviewed and further edited by the authors, and all manuscript edits were reviewed and adapted before being included in the manuscript. We take responsibility for the final content of this work, including text, claims or artifacts produced with the aid of generative AI.

\subsection*{Ethics statement}

The authors do not identify any ethical concerns related to the work reported in this article.



\subsection*{Reproducibility statement}

The code and data that produce the results reported in this paper are available at \url{https://anonymous.4open.science/r/SemTAC-2C28/} and \url{https://anonymous.4open.science/r/ContrastLearn-C97C/}.
Additionally, we provide full information for environment reproducibility, both through dependency listing in the code and describing the computing environment in Appendix \ref{compute-resources}.
Details to reproduce the data synthesis can also be found in \ref{sec:dataset-preparation} and Appendix \ref{app:generation_details}.
Besides the main results presented in the Results section of the main paper, we also include the full report of statistical comparison results in Appendix \ref{result_details}.



\subsubsection*{Acknowledgments}
This work has been funded by the Deutsche Forschungsgemeinschaft (DFG, German Research Foundation) - 459360854 as part of the Research Unit "Lifespan AI: From Longitudinal Data to Lifespan Inference in Health" (DFG FOR 5347), University of Bremen (https://lifespanai.de/)

\bibliography{iclr2027_conference}

@inproceedings{tevet_human_2023,
	title = {Human {Motion} {Diffusion} {Model}},
	url = {https://openreview.net/pdf?id=SJ1kSyO2jwu},
	language = {en},
	urldate = {2024-07-09},
	booktitle = {The {Eleventh} {International} {Conference} on {Learning} {Representations}},
	publisher = {ICLR},
	author = {Tevet, Guy and Raab, Sigal and Gordon, Brian and Shafir, Yonatan and Cohen-Or, Daniel and Bermano, Amit H.},
	month = feb,
	year = {2023},
}

@conference{biosignals26,
    author={Lourenço Rodrigues and Markus Wenzel and Felix Putze},
    title={Controlled Large Scale Synthetic Motion Dataset Generation Leveraging Text-to-Motion and Sample-Wise Quality Assurance},
    booktitle={Proceedings of the 19th International Joint Conference on Biomedical Engineering Systems and Technologies - Volume 1: BIOSIGNALS},
    year={2026},
    pages={140-150},
    publisher={SciTePress},
    organization={INSTICC},
    doi={10.5220/0014440500004070},
    isbn={978-989-758-802-0},
}

@article{Jekel2019,
author = {Jekel, Charles F and Venter, Gerhard and Venter, Martin P and Stander, Nielen and Haftka, Raphael T},
doi = {10.1007/s12289-018-1421-8},
issn = {1960-6214},
journal = {International Journal of Material Forming},
month = {may},
title = {{Similarity measures for identifying material parameters from hysteresis loops using inverse analysis}},
url = {https://doi.org/10.1007/s12289-018-1421-8},
year = {2019}
}

@inproceedings{Richardson2025,
author = {Richardson, Anthony and Putze, Felix},
title = {Motion Diffusion Autoencoders: Enabling Attribute Manipulation in Human Motion Demonstrated on Karate Techniques},
year = {2025},
isbn = {9798400714993},
publisher = {Association for Computing Machinery},
address = {New York, NY, USA},
url = {https://doi.org/10.1145/3716553.3750773},
doi = {10.1145/3716553.3750773},
booktitle = {Proceedings of the 27th International Conference on Multimodal Interaction},
pages = {372–380},
numpages = {9},
location = {
},
series = {ICMI '25}
}

@misc{oord_representation_2019,
	title = {Representation {Learning} with {Contrastive} {Predictive} {Coding}},
	url = {http://arxiv.org/abs/1807.03748},
	doi = {10.48550/arXiv.1807.03748},
	language = {en},
	urldate = {2026-04-09},
	publisher = {arXiv},
	author = {Oord, Aaron van den and Li, Yazhe and Vinyals, Oriol},
	month = jan,
	year = {2019},
	note = {arXiv:1807.03748 [cs]},
}

@inproceedings{chen_simple_2020,
	series = {{ICML}'20},
	title = {A simple framework for contrastive learning of visual representations},
	volume = {119},
	url = {https://dl.acm.org/doi/10.5555/3524938.3525087},
	urldate = {2026-04-21},
	booktitle = {Proceedings of the 37th {International} {Conference} on {Machine} {Learning}},
	publisher = {JMLR.org},
	author = {Chen, Ting and Kornblith, Simon and Norouzi, Mohammad and Hinton, Geoffrey},
	month = jul,
	year = {2020},
	pages = {1597--1607},
}

@article{he_momentum_2020,
	address = {Seattle, WA, USA},
	title = {Momentum {Contrast} for {Unsupervised} {Visual} {Representation} {Learning}},
	copyright = {https://ieeexplore.ieee.org/Xplorehelp/downloads/license-information/IEEE.html},
	url = {https://ieeexplore.ieee.org/document/9157636/},
	doi = {10.1109/CVPR42600.2020.00975},
	urldate = {2026-04-21},
	journal = {2020 IEEE/CVF Conference on Computer Vision and Pattern Recognition (CVPR)},
	publisher = {IEEE},
	author = {He, Kaiming and Fan, Haoqi and Wu, Yuxin and Xie, Saining and Girshick, Ross},
	month = jun,
	year = {2020},
	note = {Conference Name: 2020 IEEE/CVF Conference on Computer Vision and Pattern Recognition (CVPR)
ISBN: 9781728171685},
	pages = {9726--9735},
}

@inproceedings{wang_understanding_2020,
	series = {{ICML}'20},
	title = {Understanding contrastive representation learning through alignment and uniformity on the hypersphere},
	volume = {119},
	url = {https://dl.acm.org/doi/10.5555/3524938.3525859},
	urldate = {2026-04-21},
	booktitle = {Proceedings of the 37th {International} {Conference} on {Machine} {Learning}},
	publisher = {JMLR.org},
	author = {Wang, Tongzhou and Isola, Phillip},
	month = jul,
	year = {2020},
	pages = {9929--9939},
}

@inproceedings{sutskever_sequence_2014,
	address = {Cambridge, MA, USA},
	series = {{NIPS}'14},
	title = {Sequence to sequence learning with neural networks},
	volume = {2},
	urldate = {2026-04-21},
	booktitle = {Proceedings of the 28th {International} {Conference} on {Neural} {Information} {Processing} {Systems} - {Volume} 2},
	publisher = {MIT Press},
	author = {Sutskever, Ilya and Vinyals, Oriol and Le, Quoc V.},
	month = dec,
	year = {2014},
	pages = {3104--3112},
}

@article{hochreiter_long_1997,
	title = {Long {Short}-{Term} {Memory}},
	volume = {9},
	issn = {0899-7667},
	url = {https://ieeexplore.ieee.org/abstract/document/6795963},
	doi = {10.1162/neco.1997.9.8.1735},
	number = {8},
	urldate = {2026-04-21},
	journal = {Neural Computation},
	author = {Hochreiter, Sepp and Schmidhuber, Jürgen},
	month = nov,
	year = {1997},
	pages = {1735--1780},
}

@inproceedings{vaswani_attention_2017,
	address = {Red Hook, NY, USA},
	series = {{NIPS}'17},
	title = {Attention is all you need},
	isbn = {978-1-5108-6096-4},
	url = {https://dl.acm.org/doi/10.5555/3295222.3295349},
	urldate = {2026-04-21},
	booktitle = {Proceedings of the 31st {International} {Conference} on {Neural} {Information} {Processing} {Systems}},
	publisher = {Curran Associates Inc.},
	author = {Vaswani, Ashish and Shazeer, Noam and Parmar, Niki and Uszkoreit, Jakob and Jones, Llion and Gomez, Aidan N. and Kaiser, Łukasz and Polosukhin, Illia},
	month = dec,
	year = {2017},
	pages = {6000--6010},
}

@inproceedings{yang_hierarchical_2016,
	address = {San Diego, California},
	title = {Hierarchical {Attention} {Networks} for {Document} {Classification}},
	url = {https://aclanthology.org/N16-1174/},
	doi = {10.18653/v1/N16-1174},
	urldate = {2026-04-21},
	booktitle = {Proceedings of the 2016 {Conference} of the {North} {American} {Chapter} of the {Association} for {Computational} {Linguistics}: {Human} {Language} {Technologies}},
	publisher = {Association for Computational Linguistics},
	author = {Yang, Zichao and Yang, Diyi and Dyer, Chris and He, Xiaodong and Smola, Alex and Hovy, Eduard},
	editor = {Knight, Kevin and Nenkova, Ani and Rambow, Owen},
	month = jun,
	year = {2016},
	pages = {1480--1489},
}

@article{eldele_time-series_2021,
	address = {Montreal, Canada},
	title = {Time-{Series} {Representation} {Learning} via {Temporal} and {Contextual} {Contrasting}},
	url = {https://www.ijcai.org/proceedings/2021/324},
	doi = {10.24963/ijcai.2021/324},
	language = {en},
	urldate = {2026-04-21},
	journal = {Proceedings of the Thirtieth International Joint Conference on Artificial Intelligence},
	publisher = {International Joint Conferences on Artificial Intelligence Organization},
	author = {Eldele, Emadeldeen and Ragab, Mohamed and Chen, Zhenghua and Wu, Min and Kwoh, Chee Keong and Li, Xiaoli and Guan, Cuntai},
	month = aug,
	year = {2021},
	note = {Conference Name: Thirtieth International Joint Conference on Artificial Intelligence \{IJCAI-21\}
ISBN: 9780999241196},
	pages = {2352--2359},
}

@misc{tonekaboni_unsupervised_2021,
	title = {Unsupervised {Representation} {Learning} for {Time} {Series} with {Temporal} {Neighborhood} {Coding}},
	url = {http://arxiv.org/abs/2106.00750},
	doi = {10.48550/arXiv.2106.00750},
	urldate = {2026-04-21},
	publisher = {arXiv},
	author = {Tonekaboni, Sana and Eytan, Danny and Goldenberg, Anna},
	month = jun,
	year = {2021},
	note = {arXiv:2106.00750 [cs]},
}

@inproceedings{yue_ts2vec_2022,
	title = {{TS2Vec}: {Towards} {Universal} {Representation} of {Time} {Series}},
	volume = {36},
	issn = {2374-3468, 2159-5399},
	shorttitle = {{TS2Vec}},
	url = {https://ojs.aaai.org/index.php/AAAI/article/view/20881},
	doi = {10.1609/aaai.v36i8.20881},
	urldate = {2026-04-21},
	booktitle = {Proceedings of the {AAAI} {Conference} on {Artificial} {Intelligence}},
	author = {Yue, Zhihan and Wang, Yujing and Duan, Juanyong and Yang, Tianmeng and Huang, Congrui and Tong, Yunhai and Xu, Bixiong},
	month = jun,
	year = {2022},
	note = {Journal Abbreviation: AAAI},
	pages = {8980--8987},
}

@article{foumani_series2vec_2024,
	title = {Series2vec: similarity-based self-supervised representation learning for time series classification},
	volume = {38},
	issn = {1573-756X},
	shorttitle = {Series2vec},
	url = {https://doi.org/10.1007/s10618-024-01043-w},
	doi = {10.1007/s10618-024-01043-w},
	language = {en},
	number = {4},
	urldate = {2026-04-21},
	journal = {Data Mining and Knowledge Discovery},
	author = {Foumani, Navid Mohammadi and Tan, Chang Wei and Webb, Geoffrey I. and Rezatofighi, Hamid and Salehi, Mahsa},
	month = jul,
	year = {2024},
	pages = {2520--2544},
}

@InProceedings{pmlr-v139-radford21a,
  title = 	 {Learning Transferable Visual Models From Natural Language Supervision},
  author =       {Radford, Alec and Kim, Jong Wook and Hallacy, Chris and Ramesh, Aditya and Goh, Gabriel and Agarwal, Sandhini and Sastry, Girish and Askell, Amanda and Mishkin, Pamela and Clark, Jack and Krueger, Gretchen and Sutskever, Ilya},
  booktitle = 	 {Proceedings of the 38th International Conference on Machine Learning},
  pages = 	 {8748--8763},
  year = 	 {2021},
  editor = 	 {Meila, Marina and Zhang, Tong},
  volume = 	 {139},
  series = 	 {Proceedings of Machine Learning Research},
  month = 	 {18--24 Jul},
  publisher =    {PMLR},
  url = 	 {https://proceedings.mlr.press/v139/radford21a.html}
}

@article{von_kugelgen_self-supervised_2021,
	title = {Self-{Supervised} {Learning} with {Data} {Augmentations} {Provably} {Isolates} {Content} from {Style}},
	journal = {35th Conference on Neural Information Processing Systems (NeurIPS 2021)},
	author = {von Kügelgen, Julius and Sharma, Yash and Gresele, Luigi and Brendel, Wieland and Schölkopf, Bernhard and Besserve, Michel and Locatello, Francesco},
	year = {2021},
}

@article{kaya_deep_2019,
	title = {Deep {Metric} {Learning}: {A} {Survey}},
	volume = {11},
	copyright = {http://creativecommons.org/licenses/by/3.0/},
	issn = {2073-8994},
	shorttitle = {Deep {Metric} {Learning}},
	url = {https://www.mdpi.com/2073-8994/11/9/1066},
	doi = {10.3390/sym11091066},
	language = {en},
	number = {9},
	urldate = {2026-04-22},
	journal = {Symmetry},
	publisher = {Multidisciplinary Digital Publishing Institute},
	author = {Kaya, Mahmut and Bilge, Hasan Şakir},
	month = sep,
	year = {2019},
	pages = {1066},
}

@article{sakoe_dynamic_1978,
	title = {Dynamic programming algorithm optimization for spoken word recognition},
	volume = {26},
	copyright = {https://ieeexplore.ieee.org/Xplorehelp/downloads/license-information/IEEE.html},
	issn = {0096-3518},
	url = {http://ieeexplore.ieee.org/document/1163055/},
	doi = {10.1109/TASSP.1978.1163055},
	language = {en},
	number = {1},
	urldate = {2026-04-22},
	journal = {IEEE Transactions on Acoustics, Speech, and Signal Processing},
	author = {Sakoe, H. and Chiba, S.},
	month = feb,
	year = {1978},
	pages = {43--49},
}

@article{alt_computing_1995,
	title = {Computing the fréchet distance between two polygonal curves},
	volume = {05},
	issn = {0218-1959},
	url = {https://www.worldscientific.com/doi/abs/10.1142/S0218195995000064},
	doi = {10.1142/S0218195995000064},
	number = {01n02},
	urldate = {2026-04-22},
	journal = {International Journal of Computational Geometry \& Applications},
	publisher = {World Scientific Publishing Co.},
	author = {Alt, Helmut and Godau, Michael},
	month = mar,
	year = {1995},
	pages = {75--91},
}

@article{moon_visualizing_2019,
	title = {Visualizing structure and transitions in high-dimensional biological data},
	volume = {37},
	copyright = {2019 The Author(s), under exclusive licence to Springer Nature America, Inc.},
	issn = {1546-1696},
	url = {https://www.nature.com/articles/s41587-019-0336-3},
	doi = {10.1038/s41587-019-0336-3},
	language = {en},
	number = {12},
	urldate = {2026-08-26},
	journal = {Nature Biotechnology},
	publisher = {Nature Publishing Group},
	author = {Moon, Kevin R. and van Dijk, David and Wang, Zheng and Gigante, Scott and Burkhardt, Daniel B. and Chen, William S. and Yim, Kristina and Elzen, Antonia van den and Hirn, Matthew J. and Coifman, Ronald R. and Ivanova, Natalia B. and Wolf, Guy and Krishnaswamy, Smita},
	month = dec,
	year = {2019},
	pages = {1482--1492},
}

@article{rousseeuw_silhouettes_1987,
	title = {Silhouettes: {A} graphical aid to the interpretation and validation of cluster analysis},
	volume = {20},
	issn = {0377-0427},
	shorttitle = {Silhouettes},
	url = {https://www.sciencedirect.com/science/article/pii/0377042787901257},
	doi = {10.1016/0377-0427(87)90125-7},
	urldate = {2026-09-16},
	journal = {Journal of Computational and Applied Mathematics},
	author = {Rousseeuw, Peter J.},
	month = nov,
	year = {1987},
	pages = {53--65},
}

@article{luong_longsil_2019,
	title = {{longSil}: an {Evaluation} {Metric} to {Assess} {Quality} of {Clustering} {Longitudinal} {Clinical} {Data}},
	volume = {3},
	issn = {2509-4971},
	shorttitle = {{longSil}},
	url = {https://pmc.ncbi.nlm.nih.gov/articles/PMC8982754/},
	doi = {10.1007/s41666-019-00058-z},
	number = {4},
	urldate = {2026-09-17},
	journal = {Journal of Healthcare Informatics Research},
	author = {Luong, Duc Thanh Anh and Singh, Prerna and Ramezani, Mahin and Chandola, Varun},
	month = nov,
	year = {2019},
	pages = {441--459},
}

@misc{wu_deep_2024,
	title = {Deep {Representation} {Learning}-{Based} {Dynamic} {Trajectory} {Phenotyping} for {Acute} {Respiratory} {Failure} in {Medical} {Intensive} {Care} {Units}},
	url = {http://arxiv.org/abs/2405.02563},
	doi = {10.48550/arXiv.2405.02563},
	urldate = {2026-09-17},
	publisher = {arXiv},
	author = {Wu, Alan and Choudhary, Tilendra and Upadhyaya, Pulakesh and Ali, Ayman and Yang, Philip and Kamaleswaran, Rishikesan},
	month = may,
	year = {2024},
	note = {arXiv:2405.02563 [eess.SP]},
}

@article{facco_estimating_2017,
	title = {Estimating the intrinsic dimension of datasets by a minimal neighborhood information},
	volume = {7},
	issn = {2045-2322},
	url = {https://www.nature.com/articles/s41598-017-11873-y},
	doi = {10.1038/s41598-017-11873-y},
	language = {en},
	number = {1},
	urldate = {2026-09-02},
	journal = {Scientific Reports},
	author = {Facco, Elena and d’Errico, Maria and Rodriguez, Alex and Laio, Alessandro},
	month = sep,
	year = {2017},
	pages = {12140},
}

@book{box_time_2015,
  author    = {Box, George E. P. and Jenkins, Gwilym M. and Reinsel, Gregory C. and Ljung, Greta M.},
  title     = {Time Series Analysis: Forecasting and Control},
  edition   = {5th},
  publisher = {Wiley},
  year      = {2015},
  pages     = {181-182}
}

@book{vershynin_high-dimensional_2018,
  author    = {Vershynin, Roman},
  title     = {High-Dimensional Probability: An Introduction with Applications in Data Science},
  publisher = {Cambridge University Press},
  year      = {2018}
}
\bibliographystyle{iclr2027_conference}

\appendix
\section{Appendix}

\subsection{Motion Data Generation Details}
\label{app:generation_details}

\textbf{Prompt rephrasing.} The 20 rephrasings of each of the 10 base prompts (see Appendix~\ref{motion_prompts} for the full list) were generated by querying ChatGPT (o3-mini), instructed to produce variations that preserved the original motion instruction.

\textbf{Synthesis parameters.} Motion instances were synthesized using MDM's \verb|sample.generate| command with random seed 42 and the \verb|humanml_enc_512_50steps| checkpoint. All other parameters were left at their default values.

\textbf{Excluded motion types.} Two of the 10 motion types were excluded from sequence creation due to low quality observed during filtering (Appendix~\ref{filtering_results}): \textit{"A person shakes an object"} and \textit{"A person dances the Hula-hoop"}.

\textbf{Lateralization label verification.} For the lateralized motion types (step right/left, wave right/left), correct labeling was additionally verified: for step instances, by measuring net translation direction; for wave instances, by measuring the variability of elbow joint angles.

\subsection{Motion Prompt Generation}
\label{motion_prompts}

The prompt used to generate the 20 rephrasings for each of the initial motion type instructions was:

\begin{quote}
I am using a human motion generation model (Motion Diffusion Model) to generate a large number of samples. To guarantee a healthy level of variability in the generated samples, I want to create different phrasings for each prompt, so that they all refer to the same motion, but are encoded in a slightly different manner that promotes variability during the generation process. for each of the following initial prompts, create 20 different phrasings. the initial prompts are: \\
1 - A person jumps. \\
2 - A person does push-ups. \\
3 - A person raises their right arm and waves. \\
4 - A person raises their left arm and waves. \\
5 - A person claps. 6 - A person does a squat. \\
7 - A person takes a step to the left. \\
8 - A person takes a step to the right. \\
9 - A person shakes an object in front of them. \\
10 - A person dances the Hula-hoop.
\end{quote}

\subsection{Dataset Filtering Results}
\label{filtering_results}

\begin{table}[ht]
  \caption{Amount of accepted instances after dataset filtering using the method from \citet{biosignals26}.}
  \label{tab:filter_results}
  \centering
  \begin{tabular}{ll}
    \hline
    Motion type & \# Accepted instances \\
    \hline
    A person jumps. & 1637\\
    A person does push-ups. & 994\\
    A person raises their right arm and waves. & 1561\\
    A person raises their left arm and waves. & 2132\\
    A person claps. & 1837\\
    A person does a squat. & 1589\\
    A person takes a step to the left. & 710\\
    A person takes a step to the right. & 1378\\
    A person shakes an object in front of them. & 144\\
    A person dances the Hula-hoop. & 111\\
    \hline
  \end{tabular}
\end{table}

\subsection{Computational Resources and Environment}
\label{compute-resources}

Model training and evaluation were executed on a Ubuntu 22.04 server. Each model was trained on a single NVIDIA RTX 6000 GPU with CUDA version 12.4.
We used python version 3.10, and pytorch 2.6.0 for model training.

\subsection{Notes on Smoothness baseline score}
\label{smoothness-baseline-explanation}

Given an embedding trajectory $e_1,\dots,e_T \in \mathbb{R}^d$, we define displacement vectors $\Delta_t = e_{t+1}-e_t$ and measure smoothness as the mean cosine similarity between consecutive displacements,
\begin{equation}
    S = \frac{1}{T-2}\sum_{t=1}^{T-2}
        \frac{\langle \Delta_t, \Delta_{t+1}\rangle}{\lVert \Delta_t\rVert\,\lVert \Delta_{t+1}\rVert}.
\end{equation}
A natural but incorrect assumption is that unstructured embeddings give $S \approx 0$ at chance level. Because $\Delta_t$ and $\Delta_{t+1}$ share the term $e_{t+1}$ with opposite sign, this is not the case even when $e_1,\dots,e_T$ are i.i.d.\ isotropic random vectors with $\operatorname{Var}(e_t) = \sigma^2 I_d$:
\begin{equation}
    \operatorname{Cov}(\Delta_t,\Delta_{t+1})
    = \operatorname{Cov}(e_{t+1},e_{t+2}) - \operatorname{Var}(e_{t+1})
      - \operatorname{Cov}(e_t,e_{t+2}) + \operatorname{Cov}(e_t,e_{t+1})
    = -\sigma^2 I_d,
\end{equation}
while $\operatorname{Var}(\Delta_t) = \operatorname{Var}(\Delta_{t+1}) = 2\sigma^2 I_d$, giving a correlation of
\begin{equation}
    \rho = \frac{-\sigma^2}{2\sigma^2} = -\frac{1}{2}.
\end{equation}
Equivalently, $\{\Delta_t\}$ is a first-differenced white-noise series, i.e.\ an MA(1) process $\Delta_t = \varepsilon_{t+1}-\varepsilon_t$ with $\theta=-1$; its theoretical lag-1 autocorrelation $\rho(1) = \theta/(1+\theta^2)$ evaluates to exactly $-1/2$, the classical overdifferencing signature of applying a first difference to an already-white series \citep{box_time_2015}. Since cosine similarity between isotropic random vectors concentrates around their underlying correlation as $d$ grows \citep{vershynin_high-dimensional_2018}, $\mathbb{E}[S] \to -\tfrac12$ for unstructured embeddings rather than $0$. We therefore use $-0.5$, not $0$, as the chance-level baseline for the smoothness metric, with values approaching $+1$ indicating a smooth, locally-linear trajectory.

\subsection{Absolute values}
\label{result_details}

The statistical tests are reported using the rank-biserial correlation ($r_{rb}$) as the effect size measure. This measure ranges from -1 to +1, where (+/-)1 indicates complete separation between the two groups (positive sign indicates that first group has larger values than the second), with zero indicating no separation.

Table \ref{tab:full-sequence-comparison} shows the comparison results when looking at full sequence performance, \textit{i.e.} the F1-score of the models predicting which sequence type each sequence is when they have access to the five motions in the sequence.
\begin{table}[htp]
    \centering
    \begin{tabular}{c|cc}
        SemTAC vs. ... & Fréchet (p-value) & DTW (p-value)\\
        \hline
        TS-TCC & +0.96* $\left(2.9\text{e -}5\right)$ & +0.50 $\left(5.7\text{e -}2\right)$\\
        TS2Vec & +0.51* $\left(3.2\text{e -}2\right)$ & -0.63* $\left(1.2\text{e -}2\right)$\\
        Self-similarity & +0.20 $\left(1.7\text{e -}1\right)$ & -0.26 $\left(1.3\text{e -}1\right)$\\
        Classification & -0.56* $\left(4.9\text{e -}3\right)$ & -0.72* $\left(4.0\text{e -}4\right)$\\
    \end{tabular}
    \caption{Statistical comparison of SemTAC's sequence type classification at full length vs all other model families using the Mann Whitney U test. Results are reported as $r_{rb}$ (p-value), and significant results are marked with a *. The p-values presented are the corrected values based on the Holm-Bonferroni correction method. Comparisons are made separately for each distance method used for inference.}
    \label{tab:full-sequence-comparison}
\end{table}

Table \ref{tab:truncated-sequence-comparison} shows the comparison results when looking at truncated (only first 4 steps) sequence performance, \textit{i.e.} the F1-score of the models predicting which sequence type each sequence is when they only have access to the section of the data with label noise.

\begin{table}[htp]
    \centering
    \begin{tabular}{c|cc}
        SemTAC vs. ... & Fréchet (p-value) & DTW (p-value)\\
        \hline
        TS-TCC & +0.79* $\left(7.0\text{e -}4\right)$ & +0.88* $\left(2.0\text{e -}4\right)$\\
        TS2Vec & +0.78* $\left(7.0\text{e -}4\right)$ & +0.74* $\left(1.4\text{e -}3\right)$\\
        Self-similarity & +0.77* $\left(8.7\text{e -}7\right)$ & +0.68* $\left(1.8\text{e -}5\right)$\\
        Classification & +0.75* $\left(2.0\text{e -}4\right)$ & +0.29 $\left(1.0\text{e -}1\right)$\\
    \end{tabular}
    \caption{Statistical comparison of SemTAC's sequence type classification at length 4 vs all other model families using Mann Whitney U test. Results are reported as $r_{rb}$ (p-value), and significant results are marked with a *. The p-values presented are the corrected values based on the Holm-Bonferroni correction method. Comparisons are made separately for each distance method used for inference.}
    \label{tab:truncated-sequence-comparison}
\end{table}

\begin{table}[htp]
\centering
\begin{tabular}{l|cc|cc}
                                              & \multicolumn{4}{c}{Sequence Branching}                              \\
                                              & \multicolumn{2}{c|}{Silhouette}  & \multicolumn{2}{c}{kNN-Consistency} \\
SemTAC vs.                               & Fréchet & DTW                   & Fréchet & DTW                        \\ \hline
\multicolumn{1}{c|}{TS-TCC}                   & +0.98* $\left(1.5\text{e -}5\right)$ & +0.94* $\left(4.0\text{e -}5\right)$ 
                                              & +0.94* $\left(3.9\text{e -}5\right)$ & +0.94* $\left(5.0\text{e -}5\right)$ \\
\multicolumn{1}{c|}{TS2Vec}                   & +0.81* $\left(3.0\text{e -}4\right)$ & +0.92* $\left(4.0\text{e -}5\right)$ 
                                              & +0.81* $\left(2.6\text{e -}4\right)$ & +0.72* $\left(1.3\text{e -}3\right)$ \\
\multicolumn{1}{c|}{Self-similarity} & +0.29* $\left(4.3\text{e -}2\right)$ & +0.36* $\left(1.5\text{e -}2\right)$ 
                                              & +0.37* $\left(1.1\text{e -}2\right)$ & +0.36* $\left(1.2\text{e -}2\right)$ \\
\multicolumn{1}{c|}{Classification}  & -0.76* $\left(6.3\text{e -}5\right)$ & -0.83* $\left(1.7\text{e -}5\right)$ 
                                              & -0.91* $\left(1.7\text{e -}6\right)$ & -0.92* $\left(1.5\text{e -}6\right)$ 
\end{tabular}
\caption{Statistical comparison of SemTAC's metrics related to sequence branching vs all other model families using Mann Whitney U test. Results are reported as $r_{rb}$ (p-value), and significant results are marked with a *. The p-values presented are the corrected values based on the Holm-Bonferroni correction method. Comparisons are made separately for each distance method used for inference.}
\label{tab:space-metric-sequence-separation-comparison}
\end{table}

\begin{table}[htp]
\centering
\begin{tabular}{l|c|c}
                                              & \multicolumn{2}{c}{Temporal Encoding}       \\
SemTAC vs.                               & Order Recoverability* & Smoothness            \\ \hline
\multicolumn{1}{c|}{TS-TCC}                   & +1.00* $\left(1.5\text{e -}5\right)$ & +1.00* $\left(9.9\text{e -}6\right)$ \\
\multicolumn{1}{c|}{TS2Vec}                   & +0.70* $\left(3.0\text{e -}3\right)$ & +1.00* $\left(9.9\text{e -}6\right)$ \\
\multicolumn{1}{c|}{Self-similarity} & +0.51* $\left(2.1\text{e -}3\right)$ & +0.90* $\left(5.2\text{e -}9\right)$ \\
\multicolumn{1}{c|}{Classification}  & +0.30 $\left(9.2\text{e -}2\right)$ & +0.74* $\left(7.5\text{e -}5\right)$                
\end{tabular}
\caption{Statistical comparison of SemTAC's metrics related to temporal encoding vs all other model families using Mann Whitney U test. Results are reported as $r_{rb}$ (p-value), and significant results are marked with a *. The p-values presented are the corrected values based on the Holm-Bonferroni correction method.
* Order recoverability results are further related to the portion of variance explained by the linear projection to make sure that high recoverability values do not stem from a coincidental infinitesimal dimension.}
\label{tab:space-metric-temporal-structure-comparison}
\end{table}

\end{document}